\pdfoutput=1
\documentclass[11pt]{article}

\usepackage{acl}

\usepackage{times}
\usepackage{latexsym}

\usepackage[T1]{fontenc}
\usepackage[utf8]{inputenc}

\usepackage{microtype}

\usepackage{amsmath}
\usepackage{amstext}
\usepackage{amssymb}
\usepackage{amsthm}
\usepackage{dsfont}
\usepackage{mathtools}
\usepackage[LGR,T1]{fontenc}
\usepackage[utf8]{inputenc}

\usepackage{tikz}
\usepackage{framed}
\usepackage{algorithm}
\usepackage{algpseudocode}

\title{Number Theory Meets Linguistics: Modelling Noun Pluralisation Across 1497 Languages
  Using 2-adic Metrics}

\author{Gregory Baker \and Diego Molla-Aliod \\
  Macquarie University \\
         4 Research Park Drive \\
         gregory.baker2@hdr.mq.edu.au \and diego.molla-aliod@mq.edu.au \\
         }

\begin{document}
\maketitle
\begin{abstract}
  A simple machine learning model of pluralisation as a linear
  regression problem minimising a $p$-adic metric substantially
  outperforms even the most robust of Euclidean-space regressors on
  languages in the Indo-European, Austronesian, Trans New-Guinea,
  Sino-Tibetan, Nilo-Saharan, Oto-Meanguean and Atlantic-Congo
  language families.  There is insufficient evidence to support
  modelling distinct noun declensions as a $p$-adic neighbourhood even
  in Indo-European languages.

  %, but
  % is very pronounced for Indo-European languages.
  % performs
  % better than
  % It is possible to do linear regression using $p$-adic metrics
  % instead of Euclidean metrics. When attempting to train a machine
  % learning model to predict plural forms of words from singular forms,
  % there is a clear (and statistically significant) difference between
  % the performance of a robust Euclidean-space regressors and the
  % performance of a regressor that operates over a $p$-adic space. This

\end{abstract}

\section{Introduction}

% There is no particular reason that we should assume that our best
% models of language should involve floating point approximations of
% points in a continuous {\it Euclidean} space. For every pair of words in
% English, should there always be a word half-way between them, as a
% Euclidean space would permit? For every attempt to model how plurals
% are formed in a language, should there be a smooth transition from
% incorrect to correct?

% If anything it would seem more reasonable that integers and
% non-Euclidean spaces would make a better foundation for modelling of
% language, but this is constrained by the need for all the other
% machinery of machine learning: derivatives, gradient descent and
% measures of distance.

In this paper, we study whether $p$-adic metrics are a useful addition to
the toolkit of computational linguistics.
%Uses for $p$-adic
%metrics have been mainly found in algebraic number theory,
%and by physicists studying string theory.

It has been known in the mathematical community since 1897 —--
although only clearly since \citep{algebraischen-zahlen} --- that
there is an unusual and unexpected family of distance metrics based on
prime numbers which can be used instead of Euclidean metrics, which
have infinitesimals (to support calculus), the triangle inequality (to
support geometry), and other useful properties all the while
maintaining mathematical consistency. They are known as the $p$-adic metrics.
\citep{gouvea_padic} provides a valuable and readable introduction to
$p$-adic analysis.

Given a prime number $p$ it is possible to define\label{padic-define}
a 1-dimensional distance function $d$ as:
\begin{equation*}
\begin{aligned}
d_{p}(r, r)&=0\\
d_{p}(r, q)&=\left\{\begin{array}{ll}
1 & \text{ if } p \nmid(r-q) \\%[5pt]
\dfrac{1}{p} d_{p}\left(\frac{r}{p}, \frac{q}{p}\right) &\text { otherwise }
\end{array}\right.
\end{aligned}
\end{equation*}

(Where $x \nmid y$ means ``$x$ does not divide $y$'')

For example, if $p = 3$ then $d_{3}(1,4) = \frac{1}{3}$ and ${d_{3}(2,83)} = \frac{1}{81}$.

In particular, if $p=2$, the authors have found that the 2-adic
distance is a surprisingly useful measure for grammar morphology
tasks. In many of the languages in this study we found that
identifying the grammar rules for pluralisation turned into a problem
of finding a linear regressor which minimised a $p$-adic metric.

\begin{figure}[t]
    \centering
\tikzset{every picture/.style={line width=0.75pt}} %set default line width to 0.75pt

\begin{tikzpicture}[x=0.75pt,y=0.75pt,yscale=-0.4,xscale=1]
%uncomment if require: \path (0,363); %set diagram left start at 0, and has height of 363

%Straight Lines [id:da0935100220476125]
\draw    (140,230) -- (318,230) (186,224.5) -- (186,235.5)(232,224.5) -- (232,235.5)(278,224.5) -- (278,235.5) ;
\draw [shift={(320,230)}, rotate = 180] [color={rgb, 255:red, 0; green, 0; blue, 0 }  ][line width=0.75]    (10.93,-3.29) .. controls (6.95,-1.4) and (3.31,-0.3) .. (0,0) .. controls (3.31,0.3) and (6.95,1.4) .. (10.93,3.29)   ;
%Straight Lines [id:da45954713458993923]
\draw    (140,230) -- (140,52) (134.5,184) -- (145.5,184)(134.5,138) -- (145.5,138)(134.5,92) -- (145.5,92) ;
\draw [shift={(140,50)}, rotate = 450] [color={rgb, 255:red, 0; green, 0; blue, 0 }  ][line width=0.75]    (10.93,-3.29) .. controls (6.95,-1.4) and (3.31,-0.3) .. (0,0) .. controls (3.31,0.3) and (6.95,1.4) .. (10.93,3.29)   ;
%Straight Lines [id:da29868259120879237]
\draw  [dash pattern={on 0.84pt off 2.51pt}]  (160,210) -- (290,80) ;
%Shape: Circle [id:dp7356891521928056]
\draw  [color={rgb, 255:red, 0; green, 0; blue, 0 }  ,draw opacity=1 ][fill={rgb, 255:red, 0; green, 0; blue, 0 }  ,fill opacity=1 ] (183,185) .. controls (183,183.9) and (183.9,183) .. (185,183) .. controls (186.1,183) and (187,183.9) .. (187,185) .. controls (187,186.1) and (186.1,187) .. (185,187) .. controls (183.9,187) and (183,186.1) .. (183,185) -- cycle ;
%Shape: Circle [id:dp01123814279028601]
\draw  [color={rgb, 255:red, 0; green, 0; blue, 0 }  ,draw opacity=1 ][fill={rgb, 255:red, 0; green, 0; blue, 0 }  ,fill opacity=1 ] (276,92) .. controls (276,90.9) and (276.9,90) .. (278,90) .. controls (279.1,90) and (280,90.9) .. (280,92) .. controls (280,93.1) and (279.1,94) .. (278,94) .. controls (276.9,94) and (276,93.1) .. (276,92) -- cycle ;
%Shape: Circle [id:dp6059745823448764]
\draw  [color={rgb, 255:red, 0; green, 0; blue, 0 }  ,draw opacity=1 ][fill={rgb, 255:red, 0; green, 0; blue, 0 }  ,fill opacity=1 ] (230,138) .. controls (230,136.9) and (230.9,136) .. (232,136) .. controls (233.1,136) and (234,136.9) .. (234,138) .. controls (234,139.1) and (233.1,140) .. (232,140) .. controls (230.9,140) and (230,139.1) .. (230,138) -- cycle ;

% Text Node
\draw (175,242) node [anchor=north west][inner sep=0.75pt]   [align=left] {cat};
% Text Node
\draw (96,175) node [anchor=north west][inner sep=0.75pt]   [align=left] {cats};
% Text Node
\draw (219,241) node [anchor=north west][inner sep=0.75pt]   [align=left] {dog};
% Text Node
\draw (265,244) node [anchor=north west][inner sep=0.75pt]   [align=left] {eye};
% Text Node
\draw (96,86) node [anchor=north west][inner sep=0.75pt]   [align=left] {eyes};
% Text Node
\draw (93,129) node [anchor=north west][inner sep=0.75pt]   [align=left] {dogs};
% Text Node
\draw (94,26) node [anchor=north west][inner sep=0.75pt]   [align=left] {\bf Plural};
% Text Node
\draw (325,220) node [anchor=north west][inner sep=0.75pt]   [align=left] {\bf Singular};
\end{tikzpicture}
\caption{Pluralisation as a linear regression problem with solution $y=2^{32}x+116$}
\label{pluralisation-linear-regression}
\end{figure}
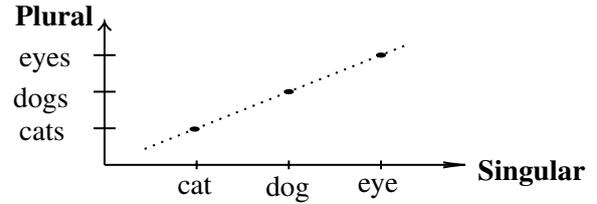

\section{Pluralisation as linear regression}\label{problem-setup}

% Finding the plural of a noun can be viewed as a machine learning
% problem, with the dual\footnote{Pun partly intended.} goal of producing
% the correct plural and also generating explanations that help a
% foreign language learner learn to perform pluralisations.

% To do so,

%First there needs to be a way of embedding words
%into a numeric vector so that machine learning algorithms can be
%used. While there are others (\citep{word-embeddings-survey} provides a
%survey) that are sophisticated,

In this paper we use a simple and naive approach for converting
vocabulary words into vectors: use whatever the unicode bit sequence
for the word would be; this bit sequence can also be viewed as an
integer vector with one element. This is of course extremely arbitrary
and subject to the whims of the unicode consortium, but it is the most
common way to represent text from any human language on a
computer.

Note that in this naive encoding scheme words like ``sky'', ``fry'' and
``butterfly'' are very close using a 2-adic metric --- the last 32 bits are
the same, meaning that the distance between them is less than or equal to than $2^{-32}$.
Using a Euclidean metric ``butterfly'' is at least
$\left(2^{32}\right)^{6}=2^{192}$ apart from the other two words. A little exploration
will observe that noun declensions in many languages --- especially
ones in the Indo-European family --- have this property that they consist of
words that form tight 2-adic clusters.

This odd correspondence between 2-adic geometry and grammar
morphology extends to declension rules for case and number where they
exist. Consider that the first two rules in Figure
\ref{english-plurals} have the property that in the naive UTF-32
encoding they can all be accurately modelled using a linear regression
performed on points in the local 2-adic neighbourhood. The fourth
rule is illustrated in Figure \ref{pluralisation-linear-regression},
with singulars and plurals of ``cat'', ``dog'' and ``eye'' plotted.
They lie on the straight line $y=2^{32}x+116$.

\begin{figure}
%  \begin{framed}
  {\small
\begin{enumerate}
\item If the singular form ends in ``y'', replace the ``y'' with ``ies''.
\item For singulars ending in ``o'' or ``i'' or ``ss'' append ``es''.
\item There are irregular nouns: ``person'' $\mapsto$ ``people'', ``sheep'' $\mapsto$ ``sheep''
\item If no other rule applies, append ``s''.
\end{enumerate}
}
%\end{framed}
\caption{A simplified and incomplete set of rules for forming plurals in English}
\label{english-plurals}
\end{figure}

\subsection{Mathematical Challenges}

Unfortunately, finding the line through a set of points that minimises
the sum of the $p$-adic measure of the residuals is harder than
finding the line that minimises the sum of the square of the
residuals. Having chosen a prime $p$, the formulation
looks similar: given a set of points
$\{ (x_i, y_i) , i \in \{ 1 \ldots N \} \}$, find $m$ and $b$ to
minimise $f(m,b) = \sum_{i=1}^N | y_i - (m x_i + b) |_p$ where
$|\cdots|_p$ is the $p$-adic measure described in section \ref{padic-define}.
But, there is no guarantee that there is a unique $(m,b)$
that minimises $f$. Consider the data set
$\{ (0,0), (1,0), (1,1), (1,2), (1,3) \}$. The 2-adic sum of distances
from those points is $\frac{5}{2}$ for $y = 0, y = x, y = 2x$ and $y = 3x$.

The derivatives of $f$ with respect to $m$ and $b$ are also
unhelpful: there are an infinite number of inflection points for any
non-trivial data set.

Fortunately, it is possible to prove that the  $p$-adic line of best fit --- unlike the
Euclidean line of best fit --- must pass through two of the data
points\footnote{In this way, the $p$-adic line of best fit is
  similar to the line of best fit supplied by the Theil-Sen, Siegel
  or RANSAC algorithms.}, which at least provides an $O(n^3)$ algorithm
for finding optimal $(m,b)$ values: draw a line through every pair of points and try them all.
The proof is in Appendix \ref{proof}.

\subsection{Data}

The dataset of singular and plural forms we used in this
research is the LEAFTOP dataset, as
described in \citep{lrec2022leaftop}. This
consists of singular and plural
noun pairs from Bible translations in 1,480 languages\footnote{
  Section \ref{experimental-results} reports results on 1,497 languages.
  In the LEAFTOP dataset,
  a language which has multiple orthographies is counted as one language
  (e.g. Chadian Arabic can also be written in a Roman alphabet), where
  in this paper each orthography has been counted as a separate language. Languages
  with significant geographic variations (such as Spanish or Portuguese)
  are also considered one language by LEAFTOP, and as multiple in this paper.}
  grouped by language family using the union of the Ethnologue \citep{ethnologue}
and Glottolog \citep{glottolog}. Since they differ on the world's primary language
families, and not every language can or should be assigned to a
language family\footnote{Klingon, for example.}, there are overlaps and gaps
in the LEAFTOP language families that are reflected in the results of
this research.

For many languages in our data set\footnote{Very little computational
  linguistics has been run on the Trans-New Guinea family of
  languages, for example.} we believe no language morphology task has
ever been run, and we thus set a baseline for these languages.

%we therefore demonstrate that $p$-adic linear
%regression is the state of the art for these.
%However, claiming a large number
%of crowns on state-of-the-art methods for different languages is not
%the aim of this research.

\section{Experiment}\label{baseline}

\begin{table}[t]
%  \begin{framed}
    {\small
\begin{tabular}{p{19mm}p{15mm}p{13mm}p{8mm}}
  \textbf{Algorithm}  & \textbf{Neigh\-bour\-hood Metric} & \textbf{Number of neighbours} & \textbf{Regr\-essor} \\
  \hline \\
{\bf Global $p$-adic} & N/A & N/A & $p$-adic \\
  {\bf Global Siegel} & N/A & N/A & Siegel \\
{\bf Local $p$-adic} & $p$-adic & 3 \ldots 20 & $p$-adic \\
{\bf Local Siegel} & Euclidean & 3 \ldots 20 & Siegel \\
  {\bf Hybrid Siegel} & $p$-adic & 3 \ldots 20 & Siegel \\
  \hline
\end{tabular}
}
\caption{Enumeration of algorithms and configurations tested, as
discussed in Section \ref{baseline}.}
\label{algorithmchoices}
%\end{framed}
\end{table}

The aim of this research
%--- apart from providing the unusual and
%somewhat amusing formulation of section \ref{problem-setup} ---
is to
identify whether or not using a $p$-adic metric space is likely to
generate improvements on computational linguistics tasks.

A linear model will obviously not be able to capture
irregular nouns. The 2-adic neighbourhood will not capture nouns that
belong to different noun declensions but share the same
ending. Comparing a linear regression model (even if it is operating
over an unusual space) to a million-parameter neural network\footnote{
  Assuming that there were computational resources and data available
  to perform this task on thousands of low-resource languages.}  where
such subtleties can be captured is going to be uninformative in
telling us about the usefulness of $p$-adic metrics.
As a result
we are comparing $p$-adic linear regression against methods that are clearly
not the state-of-the-art, but are methods which can be legitimately compared.

\begin{figure}[t]
  \begin{center}
    \includegraphics[width=0.45\textwidth]{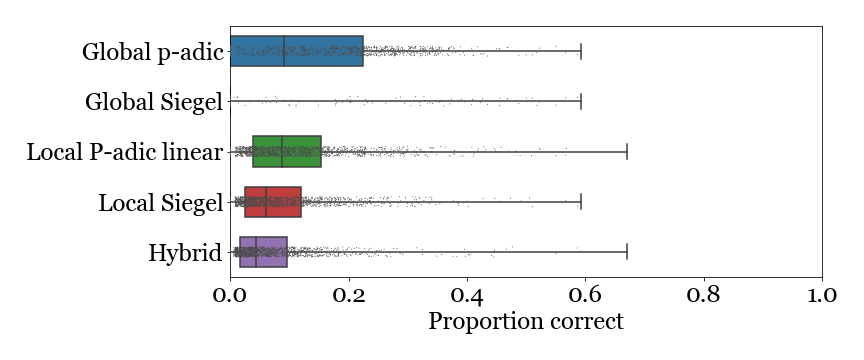}
    \end{center}
  \caption{Strip and box plot of the proportions correct for each algorithm}
  \label{all-language-correct-proportion}
\end{figure}

The choice of the Siegel regressor \citep{siegel} as the
representative for Euclidean regression was forced by the need for
robustness to a large number of outliers. The LEAFTOP data set
is known to be only 72\% accurate and any irregular nouns 
will also be outliers.
Huber \citeyear{huber}, Theil-Sen \citeyear{theil} and ordinary
least squares regression are all ruled out by these criteria.

The Siegel and $p$-adic regressors were run in ``global'' mode (learn
from as many examples as possible) and ``local'' mode (learning from a
small number of nearby words). To identify the impact of the $p$-adic
neighbourhood vs the impact of the $p$-adic linear regressor, local
Siegel was run twice, once with a $p$-adic (a ``hybrid'' of a
Euclidean regressor and a $p$-adic neighbourhood) and once with a
Euclidean neighbourhood (labeled ``local Siegel''). The complete set
of algorithms and their configurations is listed in Table
\ref{algorithmchoices}.

The only metric that can be used for this comparison is L0 ---
accuracy --- since any other metric (e.g. L1 or L2 norms) will bias
the results towards the metric space that they operate in.
A leave-one-out cross validation was done for each algorithm for each language.

\section{Results}\label{experimental-results}

A plot of results by algorithm is in Figure \ref{all-language-correct-proportion}.
Summary statistics for each language family and algorithm combination are shown
in Table \ref{heatmap-of-proportion-correct}.

\begin{table}
    \includegraphics[width=0.45\textwidth]{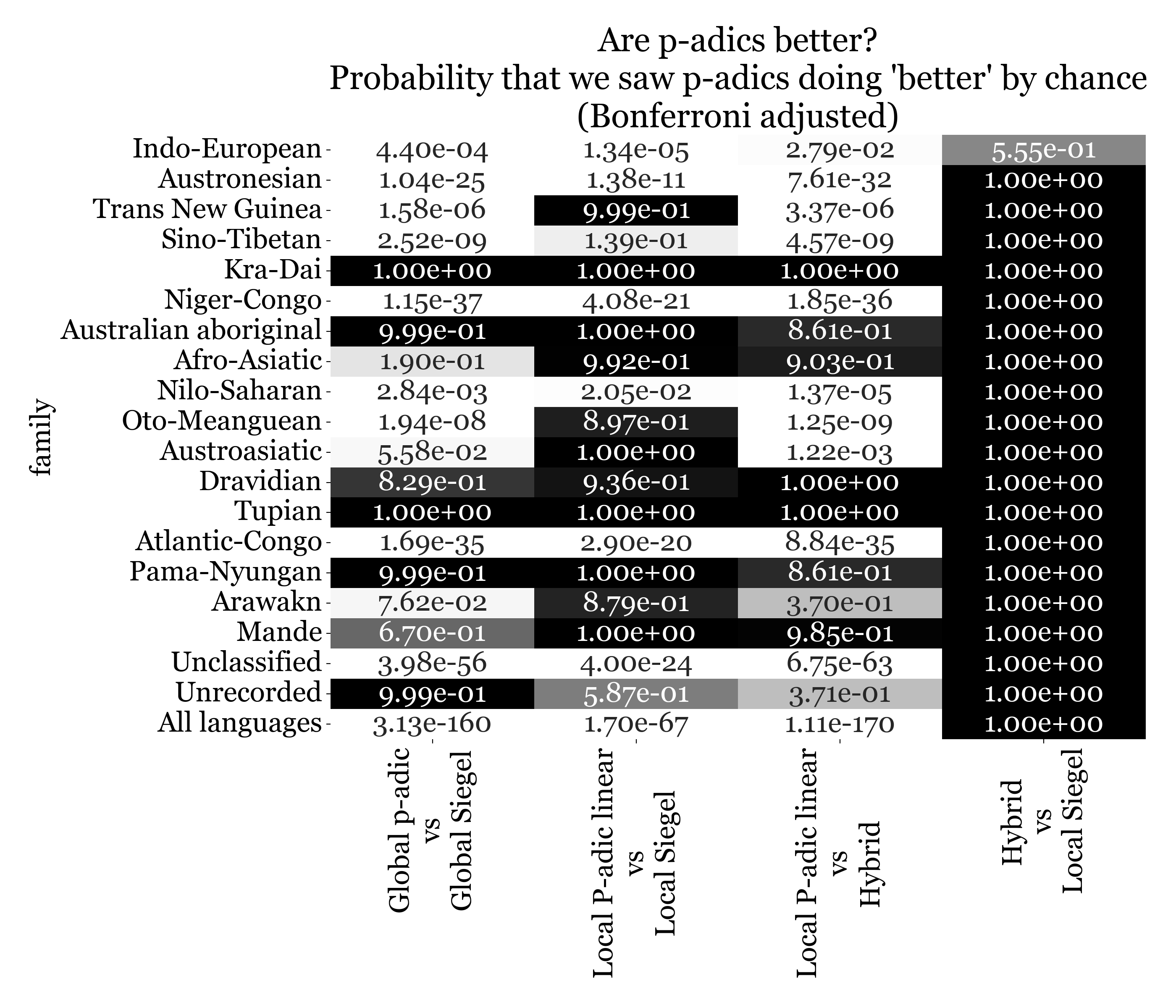}
    \caption{Experimental Results. Lighter colours indicate stronger statistical significance.}
    \label{final-experimental-results}
\end{table}

In all language families (and overall across all languages), $p$-adic
approaches outperformed Euclidean ones, however the results were not
all statistically significant.  The differences in performance between
algorithms on a language do not follow a normal distribution. Since
the research question is simply ``which is better?'' the magnitude
of the effect is unimportant, and a Wilcoxon signed-rank test can be
used. The Pratt method was used for handling situations where the
scores were identical and no sign can be calculated. The probability
is that of a one-sided result.

There are 80 statistical tests required to perform to confirm validity.
There are 17 languages families in the Ethnologue and Glottolog plus another
3 pseudo-families from the LEAFTOP labelling (Unclassifed, Unrecorded and All).
For each of these 20 families, there are 4 tests: global $p$-adic vs global Siegel;
  local $p$-adic vs local Siegel; local $p$-adic vs Siegel using a $p$-adic neighbourhood;
  Siegel with a Euclidean neighbourhood vs a $p$-adic neighbourhood.
The correction to apply to the raw statistical test results is
therefore $ p \mapsto 1 - (1-p)^{80}$.  It is this
latter (corrected) number\footnote{For example, the test result
for probability that global $p$-adic regression is equivalent to
global Euclidean Siegel on Afro-Asiatic languages is 0.00263 --- which
would have been a very clear result! --- but with 80 experiments, we
would expect to see some low-probability results. Thus the probability
of seeing a result as extreme as we saw for \textit{at least one of
  the 80 experiments} by chance is much higher: 0.23.}
that is reported in Table
\ref{final-experimental-results}.

There is strong evidence that noun pluralisation in languages in the
Indo-European, Austronesian, Trans New Guinea, Sino-Tibetan, Niger-Congo,
Nilo-Saharan, Oto-Meanguean and Atlantic-Congo families
can be modelled better with $p$-adic linear regression than
with Euclidean. This is also true for the unclassified languages
in the LEAFTOP dataset.

Moreover, the data in Table \ref{final-experimental-results}
also support the hypothesis that a randomly chosen human
language will model better using $p$-adic linear regression than
Euclidean.

\subsection{How much does a $p$-adic neighbourhood pre-filter help?}

There are many language families where training on the vocabulary in
the $p$-adic neighbourhood produced a better average correctness
score: Indo-European, Afro-Asiatic, Nilo-Saharan, Dravidian, Tupian
and Arawakan. Because of the discrepancies between the Ethnologue and
Glottolog on the categorisation of Australian languages, it appears
that there are two other language familiies (``Australian aboriginal''
and ``Pama-Nyungan'') where $p$-adic neighbourhoods are useful for
predicting the plural of a word. In addition, languages where LEAFTOP
has no language family information (``Unrecorded'') also appear to
benefit from $p$-adic neighbourhoods.

Unfortunately, none of these results hold up. The raw p-value of the
Wilcoxon test comparing global versus local $p$-adic methods on
Indo-European languages is $5.98 * 10^{-3}$,
but given that there are 9 tests to perform, the Bonferroni adjustment
tells us that the probability of seeing a result like that is $0.053$.
Close, but not compelling proof. None of the other language families
passed significance testing either.

\begin{table}[t]
  \includegraphics[width=0.5\textwidth]{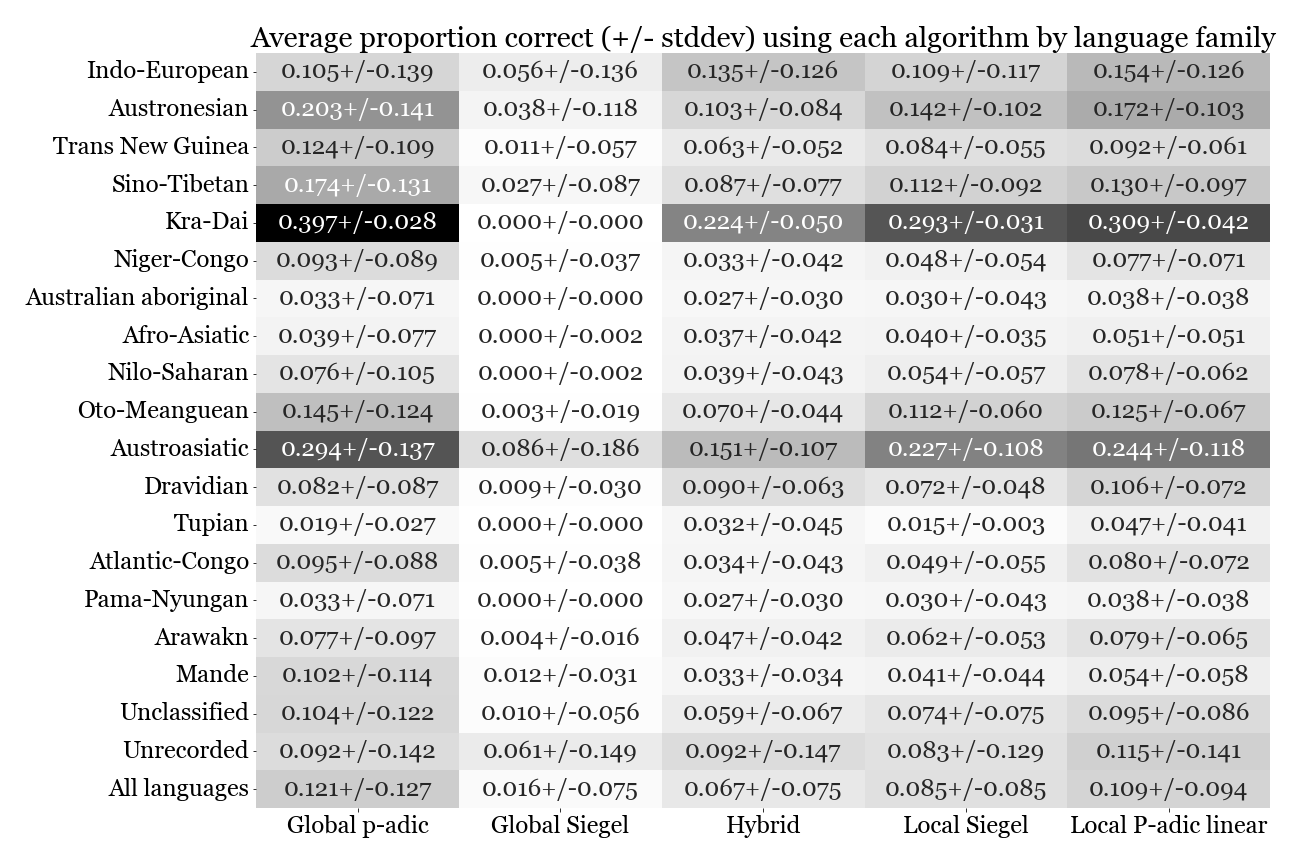}
  \caption{Average proportion correct for each combination of language family and algorithm. Darker values indicate higher accuracy.}
  \label{heatmap-of-proportion-correct}
\end{table}

Turning it around, and looking at the other 11 language families (including
``All'' and ``Unclassified''), 7 of these show a statistically significant
difference between the local and global versions of $p$-adic linear regression.
P-values for these experimental results are in Table \ref{global-vs-local}.
This can be interpreted to mean that either these language families do not
generally have noun declensions, or that using $p$-adic distance is a poor
way of separating those noun declensions.

Note also that the Hybrid algorithm (Siegel regressor trained on a
$p$-adic neighbourhood) also underperforms a Euclidean-trained Siegel
regressor.

\section{Related Work}

Murtagh (e.g. his overview paper \citealp{think-ultra}) and Bradley
(e.g. \citealp{Bradley2009OnPC}, \citealp{Bradley2008}) have written
the most on $p$-adic metrics in machine learning, having explored
clustering and support vector machines in some
depth. \citep{khrennikov-neural} provides an algorithm for training a
neural network. An extensive literature search has failed to find
any other $p$-adic adaptions of traditional machine learning algorithms.
This paper is the first to discuss $p$-adic linear regression.

Expanding the literature search more broadly, we find that there have
been very few side-by-side comparisons of Euclidean metrics versus
strongly mathematically-formulated non-Euclidean metrics for tasks in
computational linguistics.

\begin{table}[t]
  \begin{tabular}{p{28mm}p{40mm}}
    {\bf Language family} & {\bf Bonferroni-adjusted p-value of test} \\
    Austronesian & $2.39 * 10^{-6}$ \\
    Trans New Guinea & $0.032$ \\
    Sino-Tibetan & $1.92 * 10^{-5}$ \\
    Niger-Congo &  $8.76 * 10^{-7}$ \\
    Atlantic-Congo & $2.44*10^{-6}$ \\
    Unclassified & $0.0048$ \\
    All languages &  $2.69 * 10^{-13}$ \\
  \end{tabular}
  \caption{p-values of Wilcoxon tests for global $p$-adic regression versus local regression}
  \label{global-vs-local}
\end{table}

\citep{nickel2017poincare}, \citep{tifrea2018poincare} and
\citep{saxena2022cross} performed their learning of word embeddings on
a non-Euclidean metric, choosing a Poincar\'e hyperbolic
space. Calculating derivatives and finding minima of a function in a
Poincar\'e space is substantially more complex both mathematically and
computationally than for a Euclidean space. $p$-adics are simpler in
both regards, but give rise to a space with similar hyperbolic
properties. We believe that this may be a fruitful area of future
research.

\section{Conclusion}

\begin{table}[t]
  {\small
  \begin{tabular}{p{19mm}p{15mm}p{10mm}p{15mm}}
    {\bf Algorithm} & {\bf Seconds per run} & {\bf Total runs} & {\bf Approx CPU days} \\
    \hline

 Global  $p$-adic    &  8814.6 &   8643 &        881.8 \\
 Global Siegel       &    32.7 &   8643 &     3.3 \\
 Local  Siegel       &  0.368 & 155574 &     0.66 \\
 Local $p$-adic      &   10.1 & 155574 &      18.2 \\
 Hybrid Siegel       &  0.398 & 155574 &      0.72 \

  \end{tabular}
}
  \caption{Computation time}
  \label{run-time-performance}
%  \hrule
\end{table}

We demonstrated superiority over Euclidean methods on languages in the
Indo-European, Austronesian, Trans New-Guinea, Sino-Tibetan,
Nilo-Saharan and Oto-Meanguean and Atlantic-Congo language families.

Based on this, we expect that substituting $p$-adic metrics for
Euclidean metrics in other computational linguistics tasks and machine
learning methods may be an exciting area of research.

% The results showing that using a $p$-adic neighbourhood is not
% significantly helpful outside of Indo-European languages (and
% extremely border-line there) is statistical evidence
% that that noun declensions are rare.

\section*{Acknowledgements}

The authors thank the team at the Australian National Computational
Infrastructure for the grant of 170,000 hours of compute time on gadi,
without which the computations for this project could not have been
completed.

\bibliography{anthology,bibliography.bib}

\newpage
\appendix

\section{Proof  that the $p$-adic line of best fit passes through at least two points in the dataset}\label{proof}

\begin{table}
  \begin{framed}
\begin{enumerate}
\item $\forall i,j, i \ne j, \frac{y_i - y_j}{x_i - x_j} \in \mathds{Z}$
\item It contains the origin $(x_0, y_0) = (0,0)$ and one of the
  optimal lines of best fit passes through the origin, and can
  therefore be written as $y=m x$
\item $i \ne 0 \Rightarrow x_i \ne 0$
\item The data set is sorted such that $\left|\frac{y_1 - m x_1}{x_i}\right|_p \le \left|\frac{y_i - m x_i}{x_i}\right|_p$ for all $i$  where $i > 1$
\end{enumerate}
\caption{Constraints on the data set for the proof in subsection \ref{at-least-two}}
\label{heavy-constraints}
\end{framed}
\end{table}

The proof is in three sections:

\begin{enumerate}

\item A proof that a $p$-adic line of best fit must pass
  through at least one point. (Subsection \ref{at-least-one}).
\item A proof that for a data set with some strong restrictions, that
  if a $p$-adic line of best fit passes through one particular point
  in a dataset that it must pass through a second point. (Subsection
  \ref{at-least-two}).
\item A set of short proofs that every data set which doesn't
  satisfy
  those restrictions is related to a data set which does satisfy them,
  and that the $p$-adic lines of best fit can be calculated directly
  from them.
\end{enumerate}

The phrase ``optimal line'' will be used to mean ``one of the
set of lines whose $p$-adic residual sum is equal to the minimum
residual sum of any line through that data set''.

The notation $\mathrm{Res}_p(\{(x_i,y_i)\}, y=m x + b)$ will be used for
``the sum of the $p$-adic residuals of the line $y = m x + b$ on the set
$\{(x_i,y_i)\}$.

\subsection{$p$-adic best-fit lines must pass through one point}\label{at-least-one}

\begin{proof}

Suppose that there exists one or more lines that are optimal
for a given data set of size $s$, and suppose further that none of these
lines passes though any point in the data set.

Let one of these optimal lines be $y=mx+b$.

Order the points $(x_{i}, y_{i})$, in the dataset by their residuals
(smallest first) for this line:

\[
  \left|y_{i}-\hat{y}_{i}\right|_{p} \leqslant\left|y_{i+1}-\hat{y}_{i+1}\right|_{p}
\]

Since $y=mx+b$ does not pass through any point in the dataset,
$\left|\hat{y}_{0}-y_{0}\right|_{p}>0 $, and we can write the residual
$\left|\hat{y}_{0}-y_{0}\right|_{p}$
as $ap^n$ for some non-zero value of $a$ (satisfying
$\left|a\right|_p=1$) and some value (possibly zero) of $n$.
The ordering criteria means that
$\left|ap^n\right| \le  \left|y_{i}-\hat{y}_{i}\right|_{p}$ for all $i$.

Consider the line $y=mx+b-ap^n$. Its residual sum is

\begin{align*}
  &\mathrm{Res}_p(\{(x_{i}, y_{i})\}, y=mx+b-ap^n) \\
  &=  \sum_{i=0}^{s}  \left|\hat{y}_{i}-a p^{n}-y_{i}\right|_{p} \\
  &=
\left|\hat{y}_{0} - a p^{n} - y_{0}\right|_{p} +
    \sum_{i=1}^{s}  \left|\hat{y}_{i}-a p^{n}-y_{i}\right|_{p}  \\
  &= 0 + \sum_{i=1}^{s}  \left|\hat{y}_{i}-a p^{n}-y_{i}\right|_{p}  \\
  &\leq \sum_{i=1}^{s} \max( \left|\hat{y}_{i}-y_{i}\right|_{p}, \left|a p^n\right|_p) \\
  &= \sum_{i=1}^{s}  \left|\hat{y}_{i}-y_{i}\right|_{p} \\
  &< \sum_{i=0}^{s}  \left|\hat{y}_{i}-y_{i}\right|_{p} \\
  &= \mathrm{Res}_p(\{(x_{i}, y_{i})\}, y=mx+b)\\
\end{align*}

As this final line is the residual sum for the line $y=mx+b$, and
the first line is strictly less than the final, $y=mx+b-ap^n$ is
a more optimal line than $y=mx+b$, contradicting the premise.
\end{proof}

\subsection{$p$-adic best-fit lines must pass through two points}\label{at-least-two}

Consider a data set $\{(x_i,y_i)\}$ of size $s$ with the properties listed in
Table \ref{heavy-constraints}. Then the chosen optimal line which passes through the origin also
passes through another point in the dataset.

\begin{proof}
  Suppose that the chosen optimal line passes through only one point in the data set.

Let $m' = m + \frac{y_1 - m x_1}{x_1}$ and consider the residual sum of the
line $y = m' x$ (which passes through both $(x_0,y_0)$ and $(x_1,y_1)$).

\begin{align*}
  & \mathrm{Res}_p(\{(x_i,y_i)\}, y = m' x) \\
  & = \sum_{i=0}^{s} \left| (m + \frac{y_1 - m x_1}{x_1}) x_i  - y_i  \right|_p \\
  & = \left|0\right| + \left| (m + \frac{y_1 - m x_1}{x_1}) x_1 - y_i \right|_p \\
 & \phantom{\hspace{2cm}}  +  \sum_{i=2}^{s} \left| (m + \frac{y_1 - m x_1}{x_1}) x_i  - y_i  \right|_p \\
  & = \left| m x_1 + y_1 - m x_1  - y_i \right|_p \\
&  \phantom{\hspace{2cm}} + \sum_{i=2}^{s} \left| (m + \frac{y_1 - m x_1}{x_1}) x_i  - y_i  \right|_p \\
  & = 0  +  \sum_{i=2}^{s} \left| (m + \frac{y_1 - m x_1}{x_1}) x_i  - y_i  \right|_p \\
  & = \sum_{i=2}^{s} \left| m x_i - y_i + \frac{y_1 - m x_1}{x_1}) x_i  \right|_p \\
  & \le \sum_{i=2}^{s}  \max ( \left| m x_i - y_i \right|_p , \left| \frac{y_1 - m x_1}{x_1} x_i  \right|_p ) \\
  & = \sum_{i=2}^{s}  \max ( \left| m x_i - y_i \right|_p , \left| \frac{y_1 - m x_1}{x_1} \right|_p \cdot \left| x_i  \right|_p ) \\
  & \le \sum_{i=2}^{s}  \max ( \left| m x_i - y_i \right|_p , \left| \frac{y_i - m x_i}{x_i} \right|_p \cdot \left| x_i  \right|_p ) \\
  & = \sum_{i=2}^{s}  \max ( \left| m x_i - y_i \right|_p , \left| m x_i - y_i \right|_p ) \\
  & =  \sum_{i=2}^{s}  \left| m x_i - y_i \right|_p  \\
  & < 0 + \left| y_1 - m x_1 \right|_p +  \sum_{i=2}^{s}  \left| m x_i - y_i \right|_p  \\
  & = \sum_{i=0}^{s}  \left| m x_i - y_i \right|_p  \\
  & = \mathrm{Res}_p(\{(x_i,y_i)\}, y = m x)\\
\end{align*}

The last term is the residual sum from the line $y=m x$
(a line which was supposed to be optimal
for the data set), which is strictly larger
than the residual sum from $y = m' x$. This contradicts the premise.
\end{proof}

\subsection{Loosening the criteria}\label{loosen-criteria}

This subsection loosens the criteria of the proof in subsection \ref{at-least-two}.

The first three arguments (and the last half of the fourth argument)
have a common structure.

They start with a data set of points $D$ and find a way of taking an arbitrary
linear function $f$ and performing a non-singular (invertible) linear transformation
to turn them into a set $D'$ and $f'$ where the residuals of the
two functions are also invertibly linearly transformed, with the transformation
coefficients solely based on the contents of $D$.

That is, there will be a set-transformation function of the form
$T_d(x,y) = (t_0 x + t_1, t_2 y + t_3)$, a function
transformation
$T_f(f): T_f(f(x,y)) = f(t_4 x + t_5, t_6 y + t_7)$, and a residual transformation
$T_r(\mathrm{Res}_p(D,f)) = \mathrm{Res}_p(D', f') = t_{8}
\mathrm{Res}_p(D,f)) + t_{9}$.  The coefficients $t_0 \mathrm{\ldots} t_9$ are
dependent only on $D$, and $t_0, t_2, t_4, t_6 + t_8$ are all non-zero.

Thus, if a line $f$ is optimal for $D$, then the line $f'$ will be optimal for
$D'$ and vice versa. As a result, the interesting property of the optimal line $f'$ of
$D'$ (that $f'$ must pass through two points in $D'$ if it is optimal)
will also apply to $D$ and $f$.

\begin{proof}[Scaling of $y$]
Given two datasets,
$D = \{(x_i, y_i)\}$ and $D' = \{(x_i, \alpha y_i)\}$ and a line
$y=m x+b$ with a residual $r$ on $D$, there is another line
$y=\alpha m x + \alpha b$ with a residual $\left|\alpha \right|_p r$
on $D'$ (and vice versa). This is a straightforward consequence of
factorisation:

\begin{align*}
 &  \mathrm{Res}_p(\{(x_i, \alpha y_i)\}, y=\alpha m x + \alpha b) \\
  & =    \sum_i \left| \alpha m x_i + \alpha b - (\alpha y_i) \right|_p \\
    & = \left| \alpha \right|_p \cdot \sum_i \left| m x_i + b - y_i \right|_p \\
  & = \left| \alpha \right|_p \mathrm{Res}_p(\{(x_i, y_i)\}, y= m x +  b) \\
\end{align*}
\end{proof}

\begin{proof}[Scaling of $x$]
Likewise, there are relationships between data sets with scaled $x$
values. If $D = \{(x_i, y_i)\}$ and $D' = \{(\alpha x_i, y_i)\}$, then
the residual of the line $y = m x + b$ on $D$ is the same as the
residual of the line $y = \frac{m}{\alpha} x + b$ on $D'$.

\begin{align*}
 &   \mathrm{Res}_p(\{(\alpha x_i, y_i)\}, y = \frac{m}{\alpha} x + b)\\
  & = \sum_i \left| \frac{m}{\alpha} (\alpha x_i) + b - y_i \right|_p \\
  & =     \sum_i \left| m x_i + b - y_i \right|_p \\
    & =   \mathrm{Res}_p(\{(x_i, y_i)\}, y = m x + b) \\
\end{align*}
\end{proof}

Therefore, a data set having some rational (non-integer)
coefficients can be transformed into a data set with integral
coefficients where the optimal lines are similarly transformed
with only a constant multiplier effect on each residual sum simply
by multiplying through by the product of all denominators.

Moreover, if $D = \{(x_i, y_i)\}$ has integer coordinates,
then $D' = \{ \alpha x_i, y_i)\}$ where $\alpha$ is the product
$\prod_{j,k,j < k} (u_j v_k - u_k v_j)$
will not only have integer coordinates, but every line between
two points in $D'$ will have an integer gradient (and therefore
an integer y-intercept).

This generalises the result from subsection \ref{at-least-two} even when
condition (1) from Table \ref{heavy-constraints} is not satisfied.

\begin{proof}[Translation in the plane]
  Similar mechanisms apply for translation by
  a fixed offset in the $(x,y$) plane: by
adding a constant to all $x$ or $y$ values.  Given
$D = \{(x_i, y_i)\}$ and $D' = \{(x_i + a, y_i + c)\}$, the line
$y = m x + b$ has the same residual sum on $D$ as
$y = mx + (b + c - ma)$ does on $D'$.

\begin{align*}
&   \mathrm{Res}_p(\{(x_i + a, y_i + c)\}, y = mx + (b + c - ma)) \\
  & =  \sum_i \left| m (x_i + a) + (b + c - ma) - (y_i + c) \right|_p \\
  & =  \left| m x_i + b - y_i\right|_p \\
  & =    \mathrm{Res}_p(\{(x_i, y_i)\}, y = mx + b)  \\
\end{align*}
\end{proof}

This generalises the result from subsection \ref{at-least-two} to
cover data sets where condition (2) from Table \ref{heavy-constraints} is not satisfied.

\begin{proof}[When $x_i = 0$ for some or all $i$]
If condition (3)  from Table \ref{heavy-constraints} is violated, then there are two subcases to handle.

Firstly, if $x_i = 0$ for all $i$ then the optimal line is a vertical
line along the y-axis, which has the property of passing through two points in the
data set.

Alternatively, if $x_i \ne 0$ for some $i$, then define $Z$ as being
the set of points of $D$ where $x_i = 0$, and $D' = (D \setminus Z) \cup (0,0)$ where
$\setminus$ is the set difference operator.

Then for any function
$f(x)$ defined as $y = mx + b$,

\begin{align*}
  \mathrm{Res}_p(D, f) & =  \mathrm{Res}_p(D', f) + \mathrm{Res}_p(Z, f)  \\
                     & =  \mathrm{Res}_p(D', f) + \sum_{z \in Z} b - y_z \\
\end{align*}

The last term is a constant that only depends on the elements of $D$, not $f$,
thus defining an invertible linear transformation between the residuals.
\end{proof}

Condition (4) from Table \ref{heavy-constraints} can be achieved by sorting the dataset.

\section{NAACL Reproducibility Checklist}

This appendix responds to the request for reproducibility from
\citep{naacl-reproducibility}.

NAACL requirements are shown in a {\bf bold font}.

{\bf For all reported experimental results:}

\begin{itemize}
\item {\bf A clear description of the mathematical setting, algorithm, and/or model}
  Details in section \ref{problem-setup}.

\item {\bf
    A link to a downloadable source code, with specification of all dependencies, including external libraries }
  \url{https://github.com/solresol/thousand-language-morphology} and
  \url{https://github.com/solresol/padiclinear}

\item {\bf A description of computing infrastructure used} A little
  over half the computation was run on a 48-cpu node in the Gadi
  supercomputing facility. The remainder was done on Arm64 virtual machines
  running Ubuntu 21.10 at Amazon, the author's M1 Macbook Air and
  the author's x64-based Ubuntu 22.10 Linux system.

\item {\bf
    The average runtime for each model or algorithm, or estimated energy cost
  } On the author's x64-based Ubuntu system (where it was possible to
  guarantee no contention), the average run times are given in Table
  \ref{run-time-performance}.

\item {\bf
    The number of parameters in each model
  } Global P-adic and Global Siegel have no parameters. Local Siegel,
  Local P-adic Linear and Hybrid have one parameter: the number of neighbours
  to include in the training set.

\item {\bf
    Corresponding validation performance for each reported test result
  } There are not separate validation and test sets in this paper.

\item {\bf
    A clear definition of the specific evaluation measure or statistics used to report results.
  } As discussed in section \ref{baseline}, the only metric which can be used is accuracy.

\end{itemize}

{\bf
  For all results involving multiple experiments, such as hyperparameter search:
}

\begin{itemize}
\item {\bf The exact number of training and evaluation runs}
  For the Local Siegel, Local P-adic Linear and Hybrid algorithms,
  18 different neighbourhoods were explored.

\item {\bf The bounds for each hyperparameter} Minimum 3, maximum 20.
  Anything below 3 makes no sense, and with an $O(n^3)$ algorithm,
  growing beyond 20 starts to become computationally infeasible.

\item {\bf The hyperparameter configurations for best-performing models
  }
  Attached as a data file.

\item {\bf
    The method of choosing hyperparameter values (e.g. manual tuning, uniform sampling, etc.) and the criterion used to select among them (e.g. accuracy)
  } There was no need for hyperparameter selection as it was possible to
  cover the entire solution space.
\item
  {\bf
    Summary statistics of the results (e.g. mean, variance, error bars, etc.)
  }
  Detailed in section \ref{experimental-results}
\end{itemize}

{\bf Answers about all datasets used:}
See \citep{lrec2022leaftop} ---  \url{https://github.com/solresol/leaftop}

%
%\section{Example Appendix}
%\label{sec:appendix}
%
%This is an appendix.

\end{document}